\pdfoutput=1
\PassOptionsToPackage{dvipsnames}{xcolor}
\documentclass[11pt]{article}

\usepackage[final]{acl}

\usepackage{times}
\usepackage{latexsym}
\usepackage{paralist}
\usepackage{booktabs}
\usepackage{geometry}
\usepackage{multirow}
\usepackage{xcolor}
\usepackage{amsmath}
\usepackage{paralist}
\usepackage{colortbl}
\usepackage{anyfontsize}
\usepackage{graphicx}
\usepackage[T1]{fontenc}

\usepackage[utf8]{inputenc}

\usepackage{microtype}

\usepackage{inconsolata}
\newcommand{\sysname}[1]{\textsc{CrowdCounter}}

\title{\textsc{CrowdCounter}: A benchmark type-specific\\ multi-target counterspeech dataset}

\author{Punyajoy Saha\textsuperscript{\rm 1}, Abhilash Datta\textsuperscript{\rm 1}, Abhik Jana\textsuperscript{\rm 2} \and Animesh Mukherjee\textsuperscript{\rm 1} \\
       \textsuperscript{\rm 1}Indian Institute of Technology, Kharagpur, \textsuperscript{\rm 2}Indian Institute of Technology, Bhubaneswar \\
    \texttt{punyajoys@iitkgp.ac.in,abhilashdatta8224@gmail.com} \\
    \texttt{abhikjana@iitbbs.ac.in,animeshm@cse.iitkgp.ac.in }}

\begin{document}
\maketitle
\begin{abstract}

Counterspeech presents a viable alternative to banning or suspending users for hate speech while upholding freedom of expression. However, writing effective counterspeech is challenging for moderators/users. Hence, developing suggestion tools for writing counterspeech is the need of the hour. One critical challenge in developing such a tool is the lack of quality and diversity of the responses in the existing datasets. Hence, we introduce a new dataset - \sysname{} containing 3,425 hate speech-counterspeech pairs spanning six different counterspeech types (empathy, humor, questioning, warning, shaming, contradiction), which is the first of its kind. The design of our annotation platform itself encourages annotators to write type-specific, non-redundant and high-quality counterspeech. We evaluate two frameworks for generating counterspeech responses - vanilla and type-controlled prompts - across four large language models. In terms of metrics, we evaluate the responses using relevance, diversity and quality. We observe that Flan-T5 is the best model in the vanilla framework across different models. Type-specific prompts enhance the relevance of the responses, although they might reduce the language quality. DialoGPT proves to be the best at following the instructions and generating the type-specific counterspeech accurately.

\end{abstract}

\section{Introduction}

The proliferation of hate speech and offensive language has become a significant problem in the current society~\cite{israeli-tsur-2022-free}. Efforts to moderate such content using banning and suspension are ineffective as users might shift to other platforms~\cite{russo2023spillover}. Further, banning/suspension hampers the principles of freedom of speech~\cite{ullmann2020quarantining}. Hence, social scientists are focusing on alternative forms of mitigation strategies, one of which is \textit{counterspeech}. It is a response to abusive or hateful language in the form of constructive and persuasive responses. 
While counterspeech presents itself as a viable alternative following the principles of freedom of expression, it comes with challenges. A major challenge is the onus on the moderators or the users to write a good counterspeech~\cite{chung2021empowering}.

Hence, researchers across the globe are attempting to develop NLG-based suggestion tools to help moderators craft counterspeech. One major challenge of building such tools is a good quality and diverse abusive speech-counterspeech pair dataset. Few of the past datasets use synthetically generated hate speech\cite{chung-etal-2019-conan,fanton2021human}, while others are not very diverse in terms of abusive speech targets~\cite{chung-etal-2019-conan} or types of counterspeech~\cite{qian2019benchmark}. Few of the approaches require experts~\cite{chung-etal-2019-conan,fanton2021human}, which makes them less scalable. Hence, we prepare a dataset - \sysname{} following the steps listed below.

\begin{compactitem}
    \item  We use HateXplain~\cite{mathew2021hatexplain} to collect the abusive samples which has \textbf{diverse targets} and \textbf{social media dialect}.
    \item Our crowd-based annotation platform is designed to avoid common pitfalls, which reduces the dependence on \textbf{experts}.
    \item We encourage the annotators to write a particular type of counterspeech for each hate speech. This ensures \textbf{diversity} of responses. 
\end{compactitem}

\noindent Based on this, we curate a dataset having \textbf{3425} hate speech-counterspeech pairs from \textbf{1325} unique hate speech which amounts to \textbf{2.58} counterspeech per hate speech. The dataset contains six different types of counterspeech as suggested by~\citet{benesch2014countering}.  To the best of our knowledge, this is the first benchmark for evaluating type-specific counterspeech generation across various types and targets.

Using this dataset, we built two prompting frameworks -- vanilla and type-specific prompts, for generating counterspeech using four models. In the vanilla prompt approach, we also compare two parallel hatespeech-counterspeech datasets - Gab and Reddit~\cite{qian2019benchmark}. We evaluate the generated responses using three different categories of metrics - referential, diversity and quality. We make the following observations.

\begin{compactitem}
    \item Our dataset has a higher quality in terms of \textbf{diversity}, \textbf{readability} and \textbf{quality} metrics compared to other crowd-sourced datasets - Gab and Reddit. 
    \item \textbf{Flan-T5-base} emerges as the top model in the vanilla generation - generating more relevant (meteor and gleu), better quality (gruen) and diverse responses (div, dist-2). The \textbf{Llama} models are better in terms of bleurt, while the \textbf{DialoGPT} generates counterspeech with high counter-argument quality.
    \item Type-specific generation enhances the counterspeech quality and relevance (bleurt), deteriorates the language quality, and increases toxicity. \textbf{Flan-T5-base} generates the most diverse counterspeech and has better language quality. \textbf{DialoGPT} responses follow the type to be generated more accurately in terms of precision and recall. (Examples in Appendix)
\end{compactitem}

We make our annotation framework, code and dataset public at this link\footnote{\url{https://github.com/hate-alert/CrowdCounter}} for reproducibility and future research.

\section{Related works}

Counterspeech~\cite{benesch2014countering} has been proposed as an effective mitigation strategy for hate speech~\cite{cypris2022intervening,saha2022countergedi,li-etal-2022-knowledge,zhu-bhat-2021-generate}. One of the earliest works~\cite{qian2019benchmark} collected abusive language from Gab and Reddit and asked the crowd annotators to provide the counterspeech. Few other datasets~\cite{chung-etal-2019-conan,fanton2021human} rely on expert annotations. One of the key problems of both these datasets is that the hate speech instances are generated synthetically; hence, a counterspeech generation system built on this cannot be deployed on actual social media platforms.

As highlighted by \citet{benesch2016considerations}, different strategies/types are helpful while writing an effective counterspeech. \citet{mathew2019thou} curated a dataset of counterspeech, where each instance was annotated by the type(s) they corresponded to. Another dataset~\cite{chung-etal-2019-conan} also contains types annotated along with the counterspeech provided; however, it is only limited to Islamophobic content. Recently, \citet{gupta-etal-2023-counterspeeches} re-annotated the counterspeech instances from a past work~\cite{fanton2021human} with type-specific annotation.  We had difficulty accessing the dataset for our benchmarking. The data was not available in the mentioned repository- \url{https://github.com/LCS2-IIITD/quarc-counterspeech}, and we did not receive responses to our emails requesting it. Another paper~\cite{saha-etal-2024-zero-shot} focused on creating counterspeech in zero-shot setting and tries to create type-specific counterspeech using type-specific prompts. Although this is a step in the right direction, prompt based control provides limited flexiibility. Finally, we were dismayed by not being able to retrieve the dataset and use it for our benchmarking experiments. We did not find the dataset (as claimed by the authors) in the repository associated with the paper - \url{https://github.com/LCS2-IIITD/quarc-counterspeech}; moreover, the authors did not respond to our e-mails requesting the data.

In our paper, we attempt to address the limitations of the past research and present a dataset of abusive speech-counterspeech pairs \sysname{}. The abusive speech in this dataset is naturally occurring (from either X or Gab) and is diverse in terms of the number of targets. While the counterspeech is crafted by crowd annotators, we introduced a series of techniques to avoid the pitfalls of crowd-based annotations. The annotators were tasked to craft the counterspeech instances of different types (\textit{warning of consequences}, \textit{shaming/labeling}, \textit{empathy/affiliation}, \textit{humor}, \textit{contradiction} and \textit{questions}) unlike in~\cite{gupta-etal-2023-counterspeeches} where the annotators had to label an existing counterspeech with a type thus severely limiting the expression of their own opinion.

\section{Dataset curation}
In this section, we discuss the details of how \sysname{} was curated. Specifically, we discuss how we sampled the abusive language dataset, the design of the annotation platform, the selection of annotators and the final dataset curation. We employ annotators from Amazon Mechanical Turk (https://www.mturk.com/), one of the popular annotation platforms. The following subsections provide an in-depth overview of the key steps and considerations in our dataset curation process.

\subsection{Hate speech sampling}

 In order to create an abusive speech-counterspeech pairs dataset, we first need to sample the hate speech. Since we wanted the abusive speech to represent speech from the online world, we chose one of the past datasets -- HateXplain~\cite{mathew2021hatexplain}. This dataset has abusive speech from two different platforms and targets 10 different communities like African, Islamic, etc. To collect authentic abusive speech samples, we remove all the samples considered normal by two or more annotators. This amounts to around 12k data points already labeled as abusive, i.e., hate speech or offensive. We consider only the samples from Gab, around 9k data points, since Twitter recently put strict guidelines against making their data public\footnote{\url{https://developer.twitter.com/en/developer-terms/more-on-restricted-use-cases}}. Finally, we removed all the slur heavy posts (``\textit{Nogs, jews and dykes $>>>$ how enriching	}'') having less than ten non-slur words. Slur-heavy posts have less context, discourage diversity and can be easily countered using template-based denouncing strategies. After applying these filtering conditions, we are left with 7474 samples, out of which we select around 1325 random samples for our annotation.

\subsection{Definitions}

Here, we note the definitions used in the annotation framework which includes the definitions used for identifying something as abusive, i.e., hate speech/offensive and writing counterspeech of different types. 

\subsubsection{Abusive language}

This section outlines the definitions used in the annotation framework for identifying abusive content and writing counterspeech. The authors emphasize the importance of annotators personally identifying content as abusive before writing counterspeech, as this is crucial for effective moderation. We adopt definitions from a previous study~\cite{mathew2021hatexplain} who categorize abusive content into two types:

\noindent\textbf{Hate speech}: Hate speech is a language used to express hatred toward a targeted individual or group or is intended to be derogatory, to humiliate, or to insult the members of the group, based on sensitive attributes such as race, religion, ethnic origin, sexual orientation, disability, or gender.

\noindent\textbf{Offensive speech}: Offensive speech uses profanity, strongly impolite, rude or vulgar language expressed with fighting or hurtful words to insult a targeted individual or group.

\subsubsection{Counterspeech}

Counterspeech is an expression which aims to provide a positive response to hate speech with the aim to diffuse/dilute the conversation.  In addition, counterspeech should further aim to influence the bystanders to act and the perpetrators to change their views using a counterspeech post~\cite{benesch2014countering}. Moreover, there are different recommended strategies to write a counterspeech as mentioned in the literature~\cite{mathew2019thou,benesch2016considerations,chung2021multilingual}. We summarise the strategies used in this work here (see Appendix section \ref{sec:definitions} for more details)

\begin{compactitem}
    \item \textit{Warning of consequences} - Cautioning hate speakers about potential repercussions like harm caused, online consequences, etc.
    \item \textit{Shaming} - Explicitly calling out hate speech as racist, bigoted, etc. and denouncing it.
    \item \textit{Empathy/affiliation} - Responding with a friendly, empathetic tone to de-escalate hostility.
    \item \textit{Humor} - Using humor to defuse tensions and shift the conversation dynamics.
    \item \textit{Contradiction} - Highlighting contradictions in the hate speaker's stance to discredit them.
    \item \textit{Questions} - Probing the hate speaker's sources and rationale to encourage self-reflection.
\end{compactitem}

We add the examples of each of these types in the Appendix Table ~\ref{tab:counterspeech_examples}. We further ask the annotators not to write hostile counterspeech and not to include factual counterspeech as a type since it is not a recommended strategy~\cite{benesch2016considerations}.

\subsection{Design of the annotation platform}

We developed an annotation platform which was a web page providing task descriptions, instructions, and examples. Annotators were shown ten examples of abusive speech samples. For each sample, they had to write a counterspeech of a specified required type~\cite{benesch2016considerations} if they found the sample abusive. They could additionally mark any other counterspeech types employed in their response, as one hate speech sample may warrant multiple counterspeech strategies.
Several checks were implemented to ensure quality and diversity in the collected counterspeech. A word counter check requires the response to have more than five words to avoid single-word or very short responses. An open-source grammar checker~\footnote{\url{https://languagetool.org/}} was used to verify the grammatical correctness of the counterspeech. Additionally, a similarity check was performed to prevent excessive repetition. Frequently occurring counterspeech (over ten times) were identified, and their embeddings were created using \texttt{bert-base-uncased} and indexed efficiently using FAISS~\cite{douze2024faiss}. For each new counterspeech, if its cosine similarity to a frequent response exceeded $0.95$, it was flagged as a repeated instance. If any of these three checks failed, the annotator had to re-write their counterspeech response. This rigorous annotation process and criteria aimed to collect diverse, grammatically sound, and substantive counterspeech responses, ensuring a high-quality dataset.

\subsection{Selection of annotators}

We employ annotators from Amazon Mechanical Turk (AMT)\footnote{\url{https://www.mturk.com/}} using a pilot study. We design the pilot study by collecting the hate speech-counterspeech pairs from three of the past datasets~\cite{qian2019benchmark,chung-etal-2019-conan}. An expert
selected these based on the complexity of the hate speech. We selected $10$ for such pairs for the pilot study. Each annotator had to respond with a counterspeech if (s)he thinks the post is abusive. One expert manually checks the counterspeech in terms of relevance and the presence of the type mentioned.  The expert is an experienced researcher in content moderation research, particularly experienced in counterspeech writing for a period of 5+ years. (S)he is selected if (s)he writes good counterspeech in at least $8-10$ posts. We only allow the annotators having a high approval rate ($93\%$) and approved HITS ($>1000$) to participate in this task. In this task, the annotators are paid $20$ cents if they complete the pilot task. For the main task, we selected $91$ annotators out of the $194$ who participated in the pilot study.

\subsection{Main annotation task}

From the set of $1325$ abusive samples, we select 50 samples in each batch for the main annotation task. For each sample, we choose three types of counterspeech. Each hate speech and type is shown to a different annotator, and the annotators are expected to write a counterspeech of the designated type. So, we should have three different counterspeech of three different types from three different annotators. For some of the cases, however, we did not get the annotators' responses; therefore, some of the hate speech instances have less than three responses. After completing each batch of such data, an expert checks  three samples for quality control and adds the batch to the main dataset. The quality check further removes some of the annotators who still give wrong responses in the main task. The annotator has been paid \$ 1 if they completed one HIT.

\subsection{Final dataset}

Our final dataset contains 3435 abusive speech-counterspeech pairs obtained from 1325 abusive speech. Out of the 91 users selected, 44 annotators took part in the annotations. The annotators further added additional types to 1115 of their written counterspeech. Overall, the average length of the counterspeech is 20.64 words (with standard deviation $\sigma = 10.88$). Among the types, 980 are of type warning of consequences, 853 are of type questions, 803 are of type shaming, 699 are of type contradiction, 687 are of type empathy/affiliation, and 664 are of type humor. Based on the types, we perform multi-label stratification~\cite{sechidis2011stratification} to divide this dataset into train and test sets of sizes 2147 and 1288 data points. We make sure the hate speech in the test and train sets are mutually exclusive. We note the keyword distribution and targets of the abusive speech associated with different types of counterspeech in the Appendix (Tables \ref{tab:target_info} and \ref{tab:top_keywords} respectively).

\section{Other datasets}

Here, we note the other crowd-sourced hate speech-counterspeech pairs (HS-CS) datasets that were used to compare with our dataset. We also note the curation of an additional dataset, which was used to build the multilabel type classifier (section \ref{sec:experiments}).

\subsection{HS-CS datasets}

In order to evaluate the effectiveness of \sysname{} as a benchmark dataset, we compare it with two crowd-sourced public datasets~\cite{qian2019benchmark} - Reddit and Gab that contain hate speech and its corresponding counterspeech. Reddit and Gab datasets contain $5,257$ and $14,614$ hate speech instances, respectively. We randomly take 500 hate speech samples from both these datasets and collect the corresponding counterspeeches to make the test dataset. In order to maintain size parity across all the datasets, we sampled 2000 data points and used them for training for each of these datasets. The test sizes are left intact. The details of these datasets (in terms of HS-CS pairs) are noted in Table \ref{tab:dataset_comparison_2}.

\begin{table}[!htpb]
\scriptsize
\centering
\begin{tabular}{lcc}
\toprule
\textbf{Dataset} & \textbf{\#train} & \textbf{\#test} \\
\midrule
Gab & 40106 & 1474 \\
Reddit & 12839 & 1384 \\
\sysname{} & 2147 & 1288 \\\hline
Type data & 4136 & 1018\\
\bottomrule
\end{tabular}
\caption{\footnotesize Training and testing splits for each dataset.}
\label{tab:dataset_comparison_2}
\end{table}

\subsection{Type classification dataset}
\label{sec:additional}

We use two datasets from \citet{mathew2018analyzing} and \citet{chung2021multilingual} where each counterspeech is associated with one or more types. We merge these two datasets to create a pool of 9963 samples. We remove all the samples with one label as ``\textit{hostile}'', primarily present in the dataset~\cite{mathew2018analyzing}. Finally, for each datapoint, we remove the labels which are not one of the six types that we have considered. Finally, we are left with 5154 samples. Based on the types, we perform a multi-label stratification~\cite{sechidis2011stratification} to divide the dataset into train, validation and test in the ratio of 60:20:20, respectively. We use this dataset to train a model that can classify the counterspeech type(s) given a (generated) counterspeech. We note the statistics in table \ref{tab:dataset_comparison_1}.

\section{Models}

Here, we briefly mention the models utilized in this work for counterspeech generation or counterspeech-type classification.

\noindent\textbf{BERT}~\cite{devlin-etal-2019-bert}:BERT is a pre-trained language model that has revolutionized natural language processing tasks. Developed by Google AI researchers, BERT's bidirectional training approach allows it to understand the context better, leading to improved performance~\cite{devlin-etal-2019-bert}.  We use the \texttt{bert-base-uncased}\footnote{\url{https://huggingface.co/google-bert/bert-base-uncased}} model having 110M parameters. This model is used for counterspeech-type classification.

\noindent\textbf{DialoGPT}~\cite{zhang2020dialogpt}:
DialoGPT \cite{zhang2020dialogpt} is a dialogue-centric language model developed by Microsoft, derived from the GPT-2 architecture and fine-tuned on a large dataset of Reddit conversations. It generates human-like, contextually relevant responses in multi-turn dialogues, making it well-suited for conversational AI applications like chatbots and dialogue systems. 
We use the \texttt{DialoGPT-medium}\footnote{\url{https://huggingface.co/microsoft/DialoGPT-medium}}, which has 250M parameters. This model is used for counterspeech generation.
    
\noindent\textbf{Flan-T5}~\cite{chung2022scaling}: 
FlanT5 is a large language model developed by Google that builds upon the T5 encoder-decoder architecture. It was trained on a vast and diverse corpus using a unified text-to-text framework, enabling strong performance across a wide range of natural language processing tasks. FLAN-T5's massive scale and innovative training approach have pushed the boundaries of few-shot learning, allowing it to adapt quickly to new tasks with just a few examples. We use the \texttt{flan-t5-base}\footnote{\url{https://huggingface.co/google/flan-t5-base}} having 250M parameters. This model was used for both counterspeech generation and counterspeech type classification.

\noindent\textbf{Llama}~\cite{touvron2023llama}:
Llama is a finely-tuned generative text model designed by Meta. These are trained on a diverse mix of publicly available online data between January 2023 and July 2023, and this model utilizes supervised fine-tuning (SFT) and reinforcement learning with human feedback (RLHF) to align with human preferences for helpfulness and safety. We used the \texttt{Llama-2-7b-chat-hf}\footnote{\url{https://huggingface.co/meta-llama/Llama-2-7b-chat-hf}} and the recent \texttt{Meta-Llama-3-8B-Instruct}\footnote{ \url{https://huggingface.co/meta-llama/Meta-Llama-3-8B-Instruct}} for counterspeech generation. While the former is tuned for chat-specific scenarios, the latter is better in following instructions. We use the 4-bit quantized version of these models along with LoRA~\cite{hu2021lora} to train these models.

\begin{table*}[!t]
\scriptsize
\centering
\begin{tabular}{lcccccccccc}
\toprule
\textbf{Dataset} & \textbf{\#hs} & \textbf{\#hs-cn} & \textbf{len} & \textbf{fk} ($\downarrow$) & \textbf{dc} ($\downarrow$)& \textbf{div} & \textbf{arg} & \textbf{c-arg} & \textbf{cs} & \textbf{tox ($\downarrow$)} \\
\midrule
Gab & 13678 & 41580 & 15.54 & 8.67 \color{red}{(-13\%)} & 8.55 \color{red}{(-2\%)} & 0.73 \color{red}{(-14\%)} & 0.17 \color{red}{(-19\%)} & 0.47 \color{red}{(-17\%)} & 0.48 \color{red}{(-6\%)} & 0.15\\
Reddit & 5203 & 14223 & 16.03 & 8.80 \color{red}{(-15\%)} & 8.70 \color{red}{(-4\%)} & 0.72 \color{red}{(-15\%)} & 0.17 \color{red}{(-19\%)} & 0.44 \color{red}{(-20\%)} & 0.49 \color{red}{(-4\%)} & \cellcolor{green!30}\textbf{0.14} \\
\sysname{} & 1325 & 3435 & 20.65 & \cellcolor{green!30}\textbf{7.64} & \cellcolor{green!30}\textbf{8.35} & \cellcolor{green!30}\textbf{0.85} & \cellcolor{green!30}\textbf{0.21} & \cellcolor{green!30}\textbf{0.55} & \cellcolor{green!30}\textbf{0.51} & 0.16 \\

\bottomrule
\end{tabular}
\caption{\scriptsize Comparison of dataset statistics using quality metrics like counterspeech (cs), argument (arg), counter-argument (c\_arg), toxicity (tox) scores, readability metrics -  Fleisch Kincaid (fk) and Dale Chall (dc) and semantic diversity (div).}
\label{tab:dataset_comparison_1} 
\end{table*}

\section{Metrics}

Broadly, the metrics in this paper can be divided into three parts - referential, diversity and quality metrics. Diversity and quality metrics do not require the ground truth.

\subsection{Referential metrics} 
\label{sec:referentiral}
In terms of traditional referential metrics, we use \texttt{gleu}~\cite{wu2016google} and \texttt{meteor}~\cite{banerjee2005meteor} to measure how similar the generated counterspeech are to the ground truth references. In addition, we also report two of the recent generation metrics, \texttt{bleurt}~\cite{sellam2020bleurt} and \texttt{mover-score}~\cite{zhao2019moverscore}. These metrics correlate better with human ratings than traditional metrics like \texttt{gleu} or \texttt{meteor}. 
\subsection{Quality metrics}
\label{sec:quality}

\noindent\textbf{Argument quality}: One basic characteristic of the counterspeech is that it should be argumentative. To measure this, we use the confidence score of a \texttt{roberta-base-uncased} model\footnote{\url{https://huggingface.co/chkla/roberta-argument}} fine-tuned on the argument dataset~\cite{stab-etal-2018-cross} on the generated counterspeech.\\
\noindent\textbf{Counter-argument quality}: One can say that a counterspeech should not only be an argument but, more appropriately, a counter-argument to the abusive speech. To measure this, we use the confidence score of a \texttt{bert-base-uncased}\footnote{\url{https://huggingface.co/Hate-speech-CNERG/argument-quality-bert}} model~\cite{saha2024zero} trained to identify if the reply to an argument is counter-argument or not.\\
\noindent\textbf{Counterspeech quality}: This metric is beneficial when either ground truth is absent or only a single ground truth is present, which might not be the only way to counter. We use the confidence score from a \texttt{bert-base-uncased} ~\cite{saha2024zero}\footnote{\url{https://huggingface.co/Hate-speech-CNERG/counterspeech-quality-bert}} model trained to identify something as counterspeech or not.\\
\noindent\textbf{Toxicity}: As highlighted by \citet{howard2021terror}, counterspeech should aim to diffuse the toxic language. Hence, inherently, the language of the generated response should be non-toxic. We use the HateXplain model~\cite{mathew2021hatexplain} trained on two classes -- toxic and non-toxic\footnote{\footnotesize \url{https://huggingface.co/Hate-speech-CNERG/bert-base-uncased-hatexplain-rationale-two}} to estimate toxicity of the generated response. We report the confidence in the toxic class. Higher scores in this metric correspond to a higher level of perceived toxicity.\\ 
\noindent\textbf{Readability}: Readability measures how easily and effectively a written text can be understood by its intended audience, which might determine its engagement~\cite{pancer2019readability}. We use two of the common metrics -- Fleisch Kincaid~\cite{flesch2007flesch} and Dale Chall~\cite{dale1948formula} that have been used in the previous literature and are shown to be correlated with social media engagement.\\
\noindent\textbf{GRUEN}: The GRUEN (GRammaticality, Uncertainty, and ENtailment) metric\footnote{\url{https://github.com/WanzhengZhu/GRUEN}}~\cite{zhu2020gruen} is designed to evaluate text quality by assessing four dimensions of language generation -- grammaticalilty, focus, non-redundancy and coherence. 

\subsection{Diversity metrics}
\label{sec:diversity}

Diverse responses show their linguistic expanse. It is important as the abusive language has different targets, and various counterspeech types are possible. We employ two traditional diversity metrics: \texttt{dist-2}~\cite{li-etal-2016-diversity} and \texttt{ent-2}~\cite{baheti-etal-2018-generating}. While \texttt{dist-2} measures the proportion of distinct bigrams within the generated text, \texttt{ent-2}, or bigram entropy, calculates the text's unpredictability and richness of word pairings. Finally, we also employ a semantic diversity (\texttt{div}) metric. In this metric, we first calculate the average pairwise cosine-similarity across all the generated responses and subtract this value from $1$.

\subsection{Type-classification metrics}
\label{sec:type-class}

We utilized five metrics used in the previous work~\cite{mathew2019thou} -- accuracy, precision, recall, f1-score and hamming score for evaluating the type classification. We note the description of these metrics in the Appendix. The metrics are used in two different settings. In Table \ref{tab:type_classification}, we compare the predicted output with the ground truth of the test dataset of the counterspeech type data. In Table \ref{tab:type_specific}, we try to classify the responses generated by the models. Intuitively, if the type asked to be generated is the same as the type classified, then the model can generate that type accurately. We use the Flan-T5 (base) for calculating precision and GPT-4 for calculating recall based on the results of Table~\ref{tab:type_classification}. While precision measures how accurately the model can generate the given type of counterspeech, the recall measures if the given type is one of the predicted types.

\section{Experiments}
\label{sec:experiments}

\begin{table}[!htpb]
\centering
\scriptsize
\setlength{\tabcolsep}{4pt}
\begin{tabular}{lcccccr}
\toprule
Model             & Ham. Loss ($\downarrow$)  & Accuracy & Precision & Recall & F1 Score \\ \midrule
BERT              &    0.25      &  0.27    &  0.31     & 0.27   & 0.29     \\
Flan-T5 (b)       &    \textbf{0.18}      &  \textbf{0.47}    &  \textbf{0.50}     &  0.47  &\textbf{ 0.49}     \\
GPT-4      &    0.27      &  0.37    &  0.38     &  \textbf{0.66}  & \textbf{0.49}    \\ \bottomrule
\end{tabular}
\caption{\scriptsize This table shows the comparison of different models trained and tested on the counterspeech type dataset for the task of type classification. GPT-4 is used in zero-shot setting. We use the accuracy, precision, hamming loss, recall, and F1-score.}
\label{tab:type_classification}
\end{table}
\begin{table*}[!htpb]
\scriptsize
\centering
\setlength{\tabcolsep}{4pt}
\begin{tabular}{lccccccccccccccc}
\toprule
\textbf{Model} & \textbf{gleu} & \textbf{meteor} & \textbf{bleurt} & \textbf{mover} & \textbf{nov} & \textbf{div} & \textbf{dist-2} & \textbf{ent-2} & \textbf{gruen} & \textbf{arg} & \textbf{c-arg} & \textbf{cs} & \textbf{tox ($\downarrow$)}\\\hline
\multicolumn{14}{c}{\textbf{Gab}} \\\hline
DialoGPT & 0.01 & 0.11 & -0.62 & 0.01 & 0.68 & 0.65 & 0.60 & \textbf{11.07} & 0.61 & 0.15 & \textbf{0.46} & 0.42 & 0.17 \\
Flan-T5 (b) & \textbf{0.03} & \textbf{0.18} & -0.59 & \textbf{0.08} & 0.51 & \textbf{0.69} & \textbf{0.77} & 10.84 & \textbf{0.80} & \textbf{0.17} & 0.42 & \textbf{0.50} & \textbf{0.14} \\
Llama-2 & 0.02 & 0.13 & \textbf{-0.59} & 0.04 & 0.67 & 0.62 & 0.68 & 9.87 & 0.68 & 0.08 & 0.42 & 0.22 & 0.18 \\
Llama-3 & 0.01 & 0.10 & -0.63 & 0.00 & \textbf{0.71} & 0.66 & 0.57 & 10.42 & 0.54 & 0.09 & 0.42 & 0.23 & 0.19 \\
\hline
\multicolumn{14}{c}{\textbf{Reddit}} \\\hline
DialoGPT & 0.01 & 0.10 & -0.64 & 0.01 & 0.65 & 0.68 & 0.59 & \textbf{11.32} & 0.61 & 0.14 & \textbf{0.47} & 0.32 & 0.26 \\
Flan-T5 (b) & \textbf{0.03} & \textbf{0.19} & -0.62 & \textbf{0.08} & 0.50 & \textbf{0.68} & \textbf{0.78} & 10.76 & \textbf{0.80}  & \textbf{0.17} & 0.43 & \textbf{0.44} & 0.16 \\
Llama-2 & 0.02 & 0.11 & \textbf{-0.51} & 0.04 & \textbf{0.69} & 0.60 & 0.62 & 9.72 & 0.70 & 0.11 & 0.41 & 0.24 & 0.20 \\
Llama-3 & 0.01 & 0.10 & -0.54 & 0.04 & 0.65 & 0.59 & 0.57 & 9.91 & 0.63 & 0.09 & 0.41 & 0.28 & \textbf{0.16} \\
\hline
\multicolumn{14}{c}{\textbf{\sysname{}}} \\\hline
DialogGPT & 0.01 & 0.10 & -0.75 & -0.03 & 0.70 & 0.81 & 0.59 & \textbf{11.94} & 0.67 & 0.20 & \textbf{0.53} & \textbf{0.59} & 0.15 \\
Flan-T5 (b) & \textbf{0.02} & \textbf{0.14} & -0.94 & -0.02 & 0.62 & \textbf{0.85} & \textbf{0.75} & 11.73 & \textbf{0.79} & 0.17 & 0.50 & 0.45 & \textbf{0.15}  \\
Llama-2  & 0.02 & 0.11 & -0.75 & \textbf{-0.02} & 0.78 & 0.80 & 0.61 & 11.51 & 0.67 & 0.19 & 0.51 & 0.42 & 0.21  \\
Llama-3 & 0.02 & 0.10 & \textbf{-0.67} & -0.03 & \textbf{0.80} & 0.78 & 0.55 & 11.61 & 0.64 & \textbf{0.20} & 0.49 & 0.57 & 0.18 \\
\bottomrule
\end{tabular}
\caption{\scriptsize Evaluation of vanilla responses in terms of referential, diversity and quality metrics. For evaluating referential metrics, we measure the average gleu, meteor (met), bleurt novelty (nov). For diversity, we measure average diversity (div), dist-2, ent-2.  For quality, we utilize the counterspeech (cs), argument (arg), counter-argument (c\_arg), and toxicity (tox) scores, and gruen. \textbf{Bold} denotes the best scores, and higher scores denote better performance except for toxicity.}
\label{tab:vanilla_generation}
\end{table*}

\noindent Here, we discuss our experimental setup. 

\noindent\textbf{Data statistics}: We compute different metrics to understand the quality and diversity of responses in our dataset. We compare our dataset's argument quality, counterspeech quality, toxicity, readability, and semantic diversity (div) with the Reddit and Gab datasets. For uniform comparison, we sample 3435 points from all datasets.\\
\noindent\textbf{Type classification}: To perform type classification, we use \texttt{bert-base-uncased}, \texttt{flan-t5-base} models trained on the training part of the type dataset. We use validation loss to select the best model. Hyperparameters and the instruction prompt for \texttt{flan-t5-base} are in Appendix. We also use GPT-4~\footnote{\url{https://openai.com/index/gpt-4-research/}} in a zero-shot setting on the test set. We report accuracy, precision, recall, f1-score, and hamming score.\\
\noindent\textbf{Counterspeech generation}: There are two frameworks for counterspeech generation. The \textbf{first} uses a vanilla prompt, training the model on the hate speech-counterspeech dataset from a particular dataset and testing on the same. We use 100 data points for validation and evaluate generated responses using \textit{referential}, \textit{diversity}, and \textit{quality} metrics (Table~\ref{tab:vanilla_generation}).
The \textbf{second} framework deals with type-specific counterspeech generation. We use type-specific prompts with both hate speech and counterspeech types. We train on the \sysname{} dataset, using 100 data points for validation. After training, we generate type-specific counterspeech for each hate speech and type. Hyperparameters and prompts are in Appendix. We evaluate using \textit{reference-based} and \textit{reference-free} settings. In the \textit{reference-based} setting, we select responses matching ground truth counterspeech types for each hate speech. We report type-specific response scores and changes from vanilla responses for bleurt, gruen, argument/counter-argument quality, counterspeech score, and toxicity (Table \ref{tab:type_specific_ref}). In the \textit{reference-free} setting, we use semantic diversity (div), dist-2, ent-2, gruen, argument/counter-argument quality, counterspeech score, toxicity, and precision from Flan-T5 and recall from GPT-4 considering the generated type as ground truth.

\begin{table*}[!t]
    \centering
    \setlength{\tabcolsep}{4pt}
    \scriptsize
    \begin{tabular}{rrcccccc}
    \toprule
    \textbf{Type} & \textbf{Model} & \textbf{bleurt} & \textbf{gruen} & \textbf{arg} & \textbf{c-arg} & \textbf{cs} & \textbf{tox($\downarrow$)} \\
    \midrule
    

    \multirow{4}{*}{con}  &  DialoGPT  &  -0.75 \color{ForestGreen}{(2.6\%)}  &  0.65 \color{ForestGreen}{(6.56\%)}  &  0.21 \color{red}{(-8.7\%)}  & \textbf{0.55} \color{ForestGreen}{(10.0\%)} &  0.68 \color{ForestGreen}{(9.68\%)}  & \textbf{0.15} \color{red}{(7.14\%)} \\
     &  Flan-T5 (b)  &  -0.89 \color{ForestGreen}{(7.29\%)}  & \textbf{0.74} \color{red}{(-5.13\%)} &  0.21 \color{ForestGreen}{(31.25\%)}  &  0.55 \color{ForestGreen}{(5.77\%)}  &  0.55 \color{ForestGreen}{(19.57\%)}  &  0.17 \color{red}{(30.77\%)} \ \\
     &  Llama-2  & \textbf{-0.65} \color{ForestGreen}{(14.47\%)} &  0.62 \color{red}{(-1.59\%)}  &  0.26 \color{ForestGreen}{(36.84\%)}  &  0.54 \color{ForestGreen}{(5.88\%)}  &  0.62 \color{ForestGreen}{(31.91\%)}  &  0.21 \color{ForestGreen}{(0.0\%)} \ \\
     &  Llama-3  &  -0.66 \color{ForestGreen}{(5.71\%)}  &  0.54 \color{red}{(-6.9\%)}  & \textbf{0.29} \color{ForestGreen}{(31.82\%)} &  0.5 \color{ForestGreen}{(2.04\%)}  & \textbf{0.72} \color{ForestGreen}{(16.13\%)} &  0.19 \color{red}{(26.67\%)} \ \\

        \midrule

    \multirow{4}{*}{emp}  &  DialoGPT  &  -0.69 \color{ForestGreen}{(8.0\%)}  &  0.65 \color{ForestGreen}{(6.56\%)}  & \textbf{0.22} \color{ForestGreen}{(4.76\%)} & \textbf{0.57} \color{ForestGreen}{(9.62\%)} &  0.7 \color{ForestGreen}{(20.69\%)}  &  0.17 \color{red}{(13.33\%)} \ \\
 &  Flan-T5 (b)  &  -0.77 \color{ForestGreen}{(18.09\%)}  & \textbf{0.75} \color{red}{(-3.85\%)} &  0.18 \color{red}{(-5.26\%)}  &  0.55 \color{ForestGreen}{(7.84\%)}  &  0.69 \color{ForestGreen}{(76.92\%)}  &  0.16 \color{red}{(14.29\%)} \ \\
 &  Llama-2  & \textbf{-0.54} \color{ForestGreen}{(26.03\%)} &  0.67 \color{ForestGreen}{(6.35\%)}  &  0.21 \color{ForestGreen}{(10.53\%)}  &  0.57 \color{ForestGreen}{(11.76\%)}  &  0.67 \color{ForestGreen}{(52.27\%)}  & \textbf{0.1} \color{ForestGreen}{(-54.55\%)} \\
 &  Llama-3  &  -0.59 \color{ForestGreen}{(9.23\%)}  &  0.65 \color{ForestGreen}{(10.17\%)}  &  0.2 \color{red}{(-4.76\%)}  &  0.52 \color{ForestGreen}{(10.64\%)}  & \textbf{0.73} \color{ForestGreen}{(7.35\%)} &  0.1 \color{ForestGreen}{(-37.5\%)} \ \\

\midrule

\multirow{4}{*}{hum}  &  DialoGPT  &  -0.8 \color{ForestGreen}{(3.61\%)}  &  0.65 \color{ForestGreen}{(4.84\%)}  & \textbf{0.23} \color{ForestGreen}{(15.0\%)} &  0.57 \color{ForestGreen}{(5.56\%)}  & \textbf{0.67} \color{ForestGreen}{(11.67\%)} & \textbf{0.17} \color{ForestGreen}{(0.0\%)} \\
 &  Flan-T5 (b)  &  -0.94 \color{ForestGreen}{(4.08\%)}  & \textbf{0.73} \color{red}{(-7.59\%)} &  0.18 \color{ForestGreen}{(5.88\%)}  &  0.5 \color{red}{(-1.96\%)}  &  0.61 \color{ForestGreen}{(38.64\%)}  &  0.18 \color{red}{(12.5\%)} \ \\
 &  Llama-2  &  -0.82 \color{red}{(-3.8\%)}  &  0.63 \color{ForestGreen}{(5.0\%)}  &  0.21 \color{ForestGreen}{(16.67\%)}  &  0.54 \color{ForestGreen}{(3.85\%)}  &  0.42 \color{red}{(-16.0\%)}  &  0.25 \color{red}{(47.06\%)} \ \\
 &  Llama-3  & \textbf{-0.79} \color{red}{(-9.72\%)} &  0.59 \color{ForestGreen}{(1.72\%)}  &  0.21 \color{red}{(-8.7\%)}  & \textbf{0.59} \color{ForestGreen}{(15.69\%)} &  0.58 \color{red}{(-6.45\%)}  &  0.23 \color{red}{(35.29\%)} \ \\

\midrule

\multirow{4}{*}{que}  &  DialoGPT  &  -0.76 \color{ForestGreen}{(2.56\%)}  &  0.61 \color{ForestGreen}{(0.0\%)}  & \textbf{0.17} \color{red}{(-22.73\%)} &  0.49 \color{red}{(-10.91\%)}  & \textbf{0.58} \color{ForestGreen}{(0.0\%)} &  0.23 \color{red}{(76.92\%)} \ \\
 &  Flan-T5 (b)  &  -0.99 \color{red}{(-4.21\%)}  & \textbf{0.77} \color{red}{(-1.28\%)} &  0.09 \color{red}{(-52.63\%)}  &  0.48 \color{red}{(-11.11\%)}  &  0.53 \color{ForestGreen}{(23.26\%)}  & \textbf{0.19} \color{red}{(18.75\%)} \\
 &  Llama-2  &  -0.69 \color{ForestGreen}{(8.0\%)}  &  0.6 \color{red}{(-1.64\%)}  &  0.14 \color{red}{(-22.22\%)}  &  0.52 \color{ForestGreen}{(0.0\%)}  &  0.43 \color{red}{(-15.69\%)}  &  0.25 \color{red}{(25.0\%)} \ \\
 &  Llama-3  & \textbf{-0.67} \color{ForestGreen}{(4.29\%)} &  0.52 \color{red}{(-11.86\%)}  &  0.17 \color{red}{(-22.73\%)}  & \textbf{0.53} \color{ForestGreen}{(6.0\%)} &  0.46 \color{red}{(-25.81\%)}  &  0.3 \color{red}{(87.5\%)} \ \\

\midrule

\multirow{4}{*}{sha}  &  DialoGPT  &  -0.72 \color{ForestGreen}{(2.7\%)}  &  0.63 \color{ForestGreen}{(1.61\%)}  &  0.21 \color{ForestGreen}{(10.53\%)}  &  0.52 \color{ForestGreen}{(0.0\%)}  & \textbf{0.67} \color{ForestGreen}{(11.67\%)} & \textbf{0.15} \color{red}{(7.14\%)} \\
 &  Flan-T5 (b)  &  -0.72 \color{ForestGreen}{(21.74\%)}  & \textbf{0.76} \color{red}{(-3.8\%)} &  0.2 \color{ForestGreen}{(11.11\%)}  &  0.51 \color{red}{(-1.92\%)}  &  0.63 \color{ForestGreen}{(34.04\%)}  &  0.16 \color{red}{(14.29\%)} \ \\
 &  Llama-2  & \textbf{-0.56} \color{ForestGreen}{(22.22\%)} &  0.6 \color{red}{(-3.23\%)}  &  0.22 \color{ForestGreen}{(15.79\%)}  & \textbf{0.53} \color{ForestGreen}{(6.0\%)} &  0.63 \color{ForestGreen}{(53.66\%)}  &  0.15 \color{ForestGreen}{(-28.57\%)} \ \\
 &  Llama-3  &  -0.59 \color{ForestGreen}{(6.35\%)}  &  0.49 \color{red}{(-18.33\%)}  & \textbf{0.23} \color{ForestGreen}{(9.52\%)} &  0.44 \color{red}{(-2.22\%)}  &  0.61 \color{ForestGreen}{(5.17\%)}  &  0.15 \color{ForestGreen}{(-16.67\%)} \ \\

\midrule
\multirow{4}{*}{war}  &  DialoGPT  &  -0.64 \color{ForestGreen}{(13.51\%)}  &  0.61 \color{ForestGreen}{(0.0\%)}  &  0.19 \color{red}{(-9.52\%)}  & \textbf{0.53} \color{red}{(-3.64\%)} & \textbf{0.63} \color{ForestGreen}{(8.62\%)} &  0.1 \color{ForestGreen}{(-28.57\%)} \ \\
 &  Flan-T5 (b)  &  -0.81 \color{ForestGreen}{(11.96\%)}  & \textbf{0.77} \color{red}{(-1.28\%)} &  0.16 \color{ForestGreen}{(0.0\%)}  &  0.49 \color{red}{(-7.55\%)}  &  0.37 \color{red}{(-21.28\%)}  & \textbf{0.06} \color{ForestGreen}{(-60.0\%)} \\
 &  Llama-2  & \textbf{-0.55} \color{ForestGreen}{(25.68\%)} &  0.62 \color{red}{(-1.59\%)}  &  0.17 \color{red}{(-5.56\%)}  &  0.47 \color{red}{(-9.62\%)}  &  0.52 \color{ForestGreen}{(10.64\%)}  &  0.11 \color{ForestGreen}{(-45.0\%)} \ \\
 &  Llama-3  &  -0.56 \color{ForestGreen}{(16.42\%)}  &  0.51 \color{red}{(-15.0\%)}  & \textbf{0.21} \color{ForestGreen}{(0.0\%)} &  0.46 \color{red}{(-11.54\%)}  &  0.59 \color{red}{(-6.35\%)}  &  0.06 \color{ForestGreen}{(-64.71\%)} \ \\

    \bottomrule
    \end{tabular}
    \caption{\scriptsize This table shows the evaluation of type specific responses with respect to vanilla responses for all the six categories of counterspeech. We report the type-specific scores and changes compared to vanilla generation. We measure bleurt, counterspeech (cs), argument (arg), counter-argument (c\_arg), toxicity (tox) scores and gruen. \textbf{Bold} denotes the best scores, and higher scores denote better performance except for toxicity. 
    }
    \label{tab:type_specific_ref}
\end{table*}

\begin{table*}[!t]
    \centering
    \scriptsize
    \begin{tabular}{llcccccccccc}
    \toprule
    \textbf{Type} & \textbf{Model} & \textbf{div} & \textbf{dist-2} & \textbf{ent-2} & \textbf{gruen} & \textbf{arg} & \textbf{c-arg} & \textbf{cs} & \textbf{tox($\downarrow$)} & \textbf{prec} & \textbf{rec} \\
    \midrule
    \multirow{4}{*}{con} & DialoGPT &  0.79 & 0.54 &  \textbf{12.04} & 0.64 & 0.22 &   0.55 & 0.67 & \textbf{0.16} &   0.04 & 0.82 \\
                         & Flan-T5  & \textbf{0.83} & \textbf{0.70} & 12.00 & \textbf{0.74 }& 0.21 & 0.54 & 0.57 & 0.19 & 0.05 & \textbf{0.87} \\
                         & Llama-2  & 0.79  & 0.55 & 11.63 & 0.62 & 0.25 & \textbf{0.57} & 0.60 & 0.22 & \textbf{0.05} & 0.70 \\
                         & Llama-3  & 0.78  & 0.50 & 11.53 & 0.54 & \textbf{0.27} & 0.52 & \textbf{0.67} & 0.18 & 0.02 & 0.79 \\
                         \midrule
    \multirow{4}{*}{aff} & DialoGPT & 0.78 & 0.54 &  \textbf{12.09} & 0.64 & \textbf{0.22} & 0.56 & \textbf{0.69} & 0.16 & \textbf{0.32} & \textbf{0.84} \\
                         & Flan-T5  & \textbf{0.80} & \textbf{0.66} & 11.62 & \textbf{0.74} & 0.16 & \textbf{0.56} & 0.65 & 0.16 & 0.20 & 0.66 \\
                         & Llama-2  & 0.71 & 0.56 &  10.99 & 0.67 & 0.21 & 0.53 & 0.67 & \textbf{0.13} & 0.26 & 0.50 \\
                         & Llama-3  &  0.71 & 0.54 & 11.23 & 0.64 & 0.19 & 0.52 & 0.68 & 0.11 &  0.11 &  0.37 \\
                        \midrule
    \multirow{4}{*}{hum} & DialoGPT & 0.79 & 0.54 & \textbf{12.14} & 0.64 & 0.21 & 0.55 & \textbf{0.67} & \textbf{0.16} & 0.28 & 0.13 \\
                         & Flan-T5  & \textbf{0.85} & \textbf{0.70} & 12.10 & \textbf{0.74} & 0.18 & 0.53 & 0.60 & 0.17 & \textbf{0.31} & \textbf{0.17} \\
                         & Llama-2 &   0.83 & 0.58 &  11.85 & 0.62 & 0.21 & 0.53 & 0.47 & 0.26 &  0.28 & 0.13 \\
                         & Llama-3 &   0.82 & 0.53 &  11.96 & 0.57 & \textbf{0.22} & \textbf{0.56} & 0.58 & 0.21 &  0.25 & 0.06 \\
                        \midrule
    \multirow{4}{*}{que} &  DialoGPT & 0.83 & 0.53 & \textbf{11.84} & 0.59 & 0.16 & 0.51 &\textbf{0.56} & 0.22 & \textbf{0.92} & \textbf{0.96} \\
                         &  Flan-T5  & 0.83 & \textbf{0.76} & 11.37 & \textbf{0.77} & 0.08 & 0.48 & 0.51 & \textbf{0.19} & 0.81 & 0.91 \\
                         &  Llama-2  & \textbf{0.85} & 0.54 &  11.56 & 0.60 & 0.15 & \textbf{0.52} & 0.42 & 0.27 & 0.84 &  0.95 \\
                         &  Llama-3  & 0.83 & 0.50 &  11.48 & 0.52 & \textbf{0.16} & 0.49 & 0.46 & 0.29 & 0.68 &  0.89 \\
                         
                         \midrule
    \multirow{4}{*}{sha} &  DialoGPT & \textbf{0.78} & 0.55 & \textbf{12.02} & 0.64 & 0.22 & \textbf{0.57} & \textbf{0.68} & 0.15 &  0.00 & \textbf{0.43} \\
                         &  Flan-T5  & 0.75 & \textbf{0.70} & 11.39 & \textbf{0.76} & 0.21 & 0.53 & 0.68 & \textbf{0.13} & 0.00 & 0.38 \\
                         & Llama-2   & 0.72 & 0.54 & 11.17 &  0.60 & 0.23 & 0.53 & 0.64 & 0.17 &  0.00 & 0.27  \\
                         & Llama-3   & 0.71 & 0.50 & 11.04 &  0.51 & \textbf{0.25} & 0.47 & 0.66 & 0.16 &  0.00 & 0.30 \\
                        \midrule
    \multirow{4}{*}{war} & DialoGPT & 0.71 &  0.53 &  \textbf{10.83} & 0.61 & 0.19 & \textbf{0.54} & \textbf{0.62} & 0.10 & \textbf{0.94} & \textbf{0.99} \\
                         & Flan-T5  & \textbf{0.70} & \textbf{0.78} & 9.26 & \textbf{0.78} & 0.17 & 0.49 & 0.38 & \textbf{0.06} & 0.85 & 0.98 \\
                         & Llama-2  & 0.63 & 0.56 & 10.62 & 0.62 & 0.15 & 0.48 & 0.52 & 0.11 &  0.89 & 0.97 \\
                         & Llama-3  & 0.59 & 0.49 & 9.91 &  0.49 & \textbf{0.20} & 0.46 & 0.61 & 0.06 &  0.76 &  0.92 \\
                         
    \bottomrule
    \end{tabular}
    \caption{\scriptsize This table shows the evaluation of type specific responses. We measure semantic diversity (div), dist-2, ent-2, counterspeech (cs), argument (arg), counter argument (c\_arg), toxicity (tox) scores, gruen, precision (prec) using Flan-T5 and recall (rec) using GPT-4. \textbf{Bold} denotes the best scores, and higher scores denote better performance except for toxicity.}
    \label{tab:type_specific}
\end{table*}

\section{Results}

\noindent\textbf{Comparison among datasets}: We find that \sysname{} has a higher average length of counterspeech and readability than Reddit and Gab datasets. Due to the mandatory type requirement, \sysname{} also has a higher diversity of counterspeech. \sysname{} scores higher on argument, counter-argument quality, and counterspeech quality. While toxicity is slightly higher, it is overall comparable. Table \ref{tab:dataset_comparison_1} demonstrates \sysname{}'s superiority as a counterspeech benchmark.\\
\noindent\textbf{Type classification}: For type classification, Flan-T5 has the highest performance for hamming loss, accuracy, and precision, while GPT-4 has the highest recall (Table~\ref{tab:type_classification}). BERT is the worst performer. We use Flan-T5 predictions for precision and GPT-4 for recall when evaluating generated responses (Table \ref{tab:type_specific}).\\
\noindent\textbf{Vanilla generation}: Across datasets and metrics (referential, diversity, quality in Table \ref{tab:vanilla_generation}), Flan-T5 performs best for meteor, mover's score, div, \texttt{dist-2}, and gruen. Llama models are better for bleurt and generating novel counterspeech. DialoGPT excels in counter-argument quality and \texttt{ent-2} while having low counterspeech scores for Reddit and Gab.\\
\noindent\textbf{Type-specific generation}: For \textbf{reference-based} metrics (Table~\ref{tab:type_specific_ref}), bleurt improves for most types except humor for Llama models. Language quality decreases except for DialoGPT's contradiction. Counterspeech quality improves for contradiction, empathy, and shaming. Toxicity increases for contradiction, humor, and questions but decreases for empathy and shaming. If we further compare the performances of different models across types, we find that the Llama models produce better bleurt scores, hence generating more relevant counterspeech.

For \textbf{reference-free} metrics (Table~\ref{tab:type_specific}), Flan-T5 has the best semantic diversity (div), \texttt{dist-2}, gruen, and precision. DialoGPT excels in \texttt{ent-2}. Llama-3 is best for argument quality except for empathy-affiliation. DialoGPT has the highest precision for questions and warning-of-consequences types.  In terms of recall, DialoGPT has again the highest scores for empathy-affiliation, questions, shaming and warning-of-consequences. The Llama family models are less diverse which might highlight the issue of size vs steerability for such subjective tasks. Overall, we find that no model outperforms in all counterspeech metrics. One can choose Llama for relevancy, Llama/DialoGPT for high counterspeech scores, or Flan-T5 for language quality. Further research is needed to develop a more comprehensive solution.\\
\begin{table}[!htpb]
\centering
\scriptsize
\begin{tabular}{lcc}\toprule
\textbf{Type} & \textbf{Bleurt} & \textbf{CS-score} \\\midrule
con & 0.66 & 0.66 \\
aff & 0.31 & 0.52 \\
hum & 0.18 & 0.77 \\
que & 0.17 & 0.77 \\
sha & 0.71 & 0.58 \\
war & 0.37 & 0.52 \\\bottomrule
\end{tabular}
\caption{ This table shows the Pearson's correlation between the bleurt/cs-scores and the human judgement ratings.}
\label{tab:human_correlation}
\end{table}
\noindent\textbf{Human judgement}: We took 10 generated counterspeeches each with best and worst bleurt scores for each type thus making a set of 120 samples and got them annotated by 4 experts who have long experience of research and publications on this topic. Each annotator rated each generated counterspeech on a scale of 1-5 with 1 being the worst and 5 being the best. We did the exact same exercise for 10 generated counterspeeches, but now, each with best and worst cs-scores. We measure the Pearson's correlation between the bleurt/cs-scores and the human judgement ratings. The results from these evaluations are presented in Table \ref{tab:human_correlation}. Not surprisingly we observe (as was also observed in \cite{saha-etal-2024-zero-shot}) that across all the types the correlations are positive (always $>0.5$ for at least one of the two metrics) thus reinforcing the utility of the automatic metrics we chose.

\section{Conclusion}

In conclusion, we create the first ever type-specific, diverse and crowd-sourced abusive-counterspeech pairs - \sysname{}. We trained four language models in two different frameworks i.e., vanilla and type-specific prompting. We evaluated the responses generated by these models along the dimensions of relevance, diversity and quality. We notice that compared to other crowd-sourced datasets, i.e., Gab and Reddit, \sysname{} has higher diversity and quality. In terms of vanilla generation, finetuned Flan-T5 is quite superior to even larger models from the Llama family while being $32$x smaller than them. Constraining the models to generate a particular type of counterspeech does improve the relevance of their outputs but also reduces the language quality to some extent. Finally, DialoGPT is quite proficient at following the type-specific instructions better than all the other models. Examples of generations are added in Appendix table \ref{tab:gen_counterspeech_types-1-3} and \ref{tab:gen_counterspeech_types}. Overall, this work opens up new avenues towards generating and evaluating type-specific counterspeech.

\section{Limitations}

Our work has a few limitations. Our dataset is only based on the English language, but our framework is general enough to extend to other languages as per requirement. We select the abusive content from only one specific platform - Gab, owing to various stringent policies regarding data-sharing in other platforms. Due to resource constraints, we had to run the Llama family models in quantized settings, which might have led to inferior performance compared to other models. Many of our automatic metrics are based on particular datasets, which might carry the bias of those datasets. However, we have to rely on these models to do a large-scale evaluation. 

\section{Ethics statement}

As part of data ethics, we anonymize the worker IDs before sharing the data with the public. Although our paper proposes fine-tuned large language models as counterspeech generators, we advocate against the fully automated usage of such models. We built these models as an active aid for moderators or users who wish to write counterspeech. Even then, appropriate guardrails should be applied to these models before making them public for such a sensitive task. Further, we encourage active monitoring of such counterspeech suggestion tools if deployed. 

\bibliography{custom}

\appendix

\section{Annotation details}
\label{sec:pilot-examples}

We note the pilot questions from the annotations tasks in the Table~\ref{tab:pilot_data} and the examples of the particular counterspeech in the Table~\ref{tab:counterspeech_examples}. 

\begin{figure*}[!htpb] 
    \centering
    \includegraphics[width=0.7\linewidth]{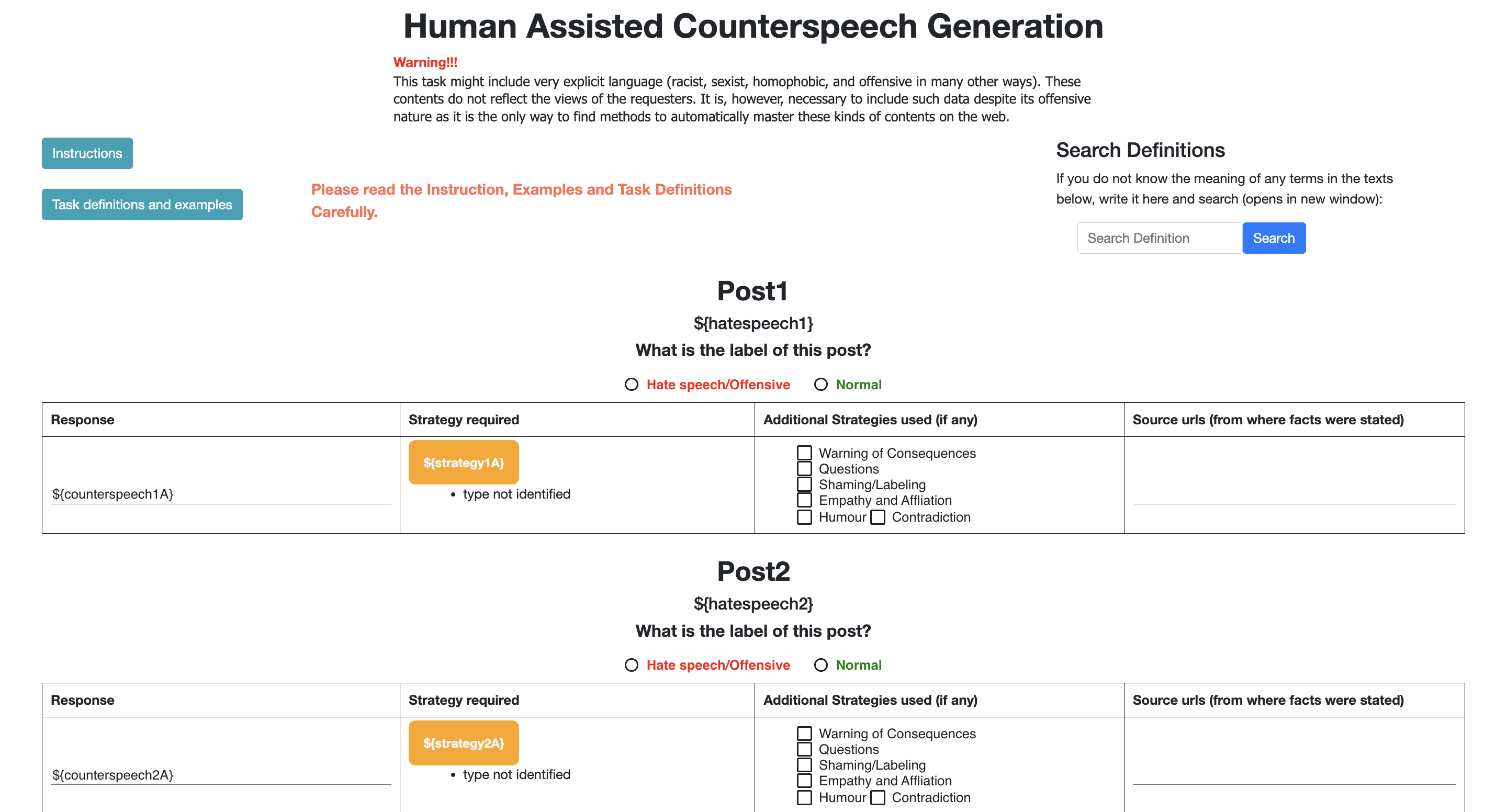} 
    \caption{The annotation platform which was used in MTurk.}
    \label{fig:annotation_platform}
\end{figure*}

\begin{table}[!htbp]
    \centering
    \scriptsize
    \begin{tabular}{p{3cm}p{3cm}l}
\toprule
\textbf{Hate speech} &  \textbf{Counterspeech} &  \textbf{Type} \\
\midrule
Minorities vote to strip wealth and power from White people and they are organised to do this by the J*ws &  Where is the evidence for this claim ? &   que \\\hline
It is enough to burn down mosques and the Koran to solve all our problems. &   That sounds a lot like terrorism to me. &             sha \\\hline
All Muslims are incompatible with any Western society. &  Millions of Muslims live in western nations. &       con \\\hline
Our national flag cannot be shown as it could annoy a Muslim.\#fitinorfuckoff., &  I am really good at making up stories, believing only the things I want to and then ranting about it too. &     hum \\\hline
We must send Islamics back to their native countries. They are too different from us. &  I have met so many Muslim people adhering to British values. What do you base your view on? & emp \\\hline
The Jews did: USS Liberty Attack, the King David hotel attack, New York 9/11, London 7/7 &  Where is the proof of this? Looks like assumptions as opposed to facts... &           que \\\hline
The world would be a better place without Muslims. They are only killing and raping our children. &  There is a problem in every community, what are you doing to stop it in yours? &       con \\\hline
The vast majority of Muslims are unemployed and refuse to work. &  All Muslim people I know are employed and hardworking individuals. Stereotyping much? & emp \\\hline
Muslim invasion is the new plague of London. They must be repatriated to their home countries! &  Invasion? Don't you think this expression is slightly over dramatic? People simply want to live in one of the best cities in the world! & sha \\\hline                                                                       
Never break bread with a Quranimal. You risk catching HIV and/or Ebola. &   If you do not think Muslims have ever done anything for 'us' I think you need to do a quick google search. The evidence of Muslims contributions to society is everywhere! &       con \\
\bottomrule
\end{tabular}
\caption{This table shows the expert-written hate speech and counterspeech pairs along with their types.}
\label{tab:pilot_data}
\end{table}

\begin{table*}[!htpb]
\centering
\footnotesize
\begin{tabular}{p{2cm}|p{6cm}|p{6cm}}
\hline
\textbf{Type} & \textbf{Hate speech} & \textbf{Counterspeech} \\ 
\hline
\vspace{1pt}
\multirow{2}{*}{\shortstack{Warning of\\consequences}} 
& The world would be a better place without Muslims. They are only killing and raping our children. 
& Online communication is permanent, so please choose your words carefully. \\ 
\cline{2-3} 
& What if our entire ruling class consisted of J-controlled paedophiles? That would explain a lot. 
& You might be banned for making such harmful and questionable claims, so please be aware. \\ 
\hline

\vspace{1pt}
\multirow{2}{*}{Shaming} 
& Islam is characterized by hate! When the first Muslim set foot in Britain, racism started in England! 
& Perhaps you should monitor your own hate speech before pointing the finger at others? \\ 
\cline{2-3} 
& Islam is all about hate! When Muslims first set foot in Britain was when race hate started in England!
& Perhaps you should monitor your own hate speech before pointing the finger at others? \\ 
\hline

\vspace{1pt}
\multirow{2}{*}{\shortstack{Empathy/\\Affiliation}} 
& I think that Muslims can be a problem for our society, because they do not want to conform to our customs. They want their religion respected, but are they respecting our laws? One day they will oppress us with the Sharia and polygamy. 
& I do not share your opinion, but I do acknowledge that opinions, mine or yours, do not have to be based on fact or knowledge. My 'experience' of Muslims, as neighbours, colleagues, friends hasn't led me to view them as a problem. \\ 
\cline{2-3} 
& \vspace{1pt} Gays are perverts. 
& Sexual orientation is a deeply personal aspect of a person's identity, and it's important to respect each individual's right to express their true selves. \\ 
\hline

\vspace{1pt}
\multirow{2}{*}{Humor} 
& Is this Afghanistan or Pakistan? You would never say this is an English town. \#saville town (text accompanying video footage of people on their way to Friday prayers etc. ).
& So these families came in to do cheap labor at textile mills in Saville and then made a life for themselves? Shocking! How could they? \\
\cline{2-3}
& Muslims in China were forced to eat pork and drink alcohol during the recent Chinese New Year celebration - At least China is doing it right!!
& Being forced to eat my peas when I was a kid just made me hate them even more. Still cannot stomach the damn things. \\
\hline

\vspace{1pt}
\multirow{2}{*}{Contradiction} 
& Children should not be taught about LGBT perversion. Families, due to their cultural or religious backgrounds, have the right to be against it.
& By this reasoning, no subject of importance should be taught in school, because there is always someone against a certain topic. \\
\cline{2-3}
& "Islam does not help science progression, because it does not have useful elements and it isn't rational."
& Apart from all the scientists that are also Muslim? \\
\hline

\vspace{1pt}
\multirow{2}{*}{Questions} 
& The world would be a better place without Muslims. They are only killing and raping our children.
& There is a problem in every community, what are you doing to stop it in yours? \\
\cline{2-3}
& What if our entire ruling class consisted of J-controlled paedophiles? That would explain a lot.
& What would it explain? This comment is very vague I wonder if you have any facts that make you wonder this. \\
\hline

\end{tabular}
\caption{Examples of counterspeech strategies that are shown to the annotators, so that they can write appropriate responses.}
\label{tab:counterspeech_examples}
\end{table*}

\begin{table} [!htpb]
    \centering
    \scriptsize
    \begin{tabular}{lrrrrrr}
\toprule
{} &  \textbf{aff} &  \textbf{hum} &  \textbf{que} &  \textbf{sha} &  \textbf{hyp} &  \textbf{war} \\
\midrule
African    &                  258 &     244 &        311 &      283 &            242 &                      358 \\
Islam      &                  189 &     162 &        235 &      218 &            209 &                      284 \\
Jewish     &                  148 &     165 &        217 &      185 &            169 &                      245 \\
Women      &                  123 &     140 &        198 &      160 &            167 &                      201 \\
Arab       &                  124 &     113 &        155 &      142 &            133 &                      184 \\
Hom  &                        98 &      98 &        129 &      131 &            100 &                      152 \\
Men        &                   73 &      72 &         99 &       81 &             88 &                      100 \\
Cau  &                         73 &      63 &         88 &       73 &             67 &                      103 \\
Refugee    &                   70 &      63 &         90 &       78 &             69 &                       88 \\
Hispanic   &                   55 &      44 &         65 &       47 &             47 &                       66 \\
\bottomrule
\end{tabular}
\caption{\scriptsize Target information of \sysname{}. The column headers refer to different types of counterspeeches -- affiliation (aff), humor (hum), questions (que), shaming (sha), hypocrisy (hyp), warning (war), and row headers refer to the targets. Abbreviated targets - Caucasian (Cau), Homosexual (Hom).}
    \label{tab:target_info}
\end{table}

\begin{table}[!htpb]
\centering
\scriptsize
\setlength{\tabcolsep}{4pt}
\begin{tabular}{lr}
\toprule
\textbf{Type} & \textbf{Top 5 keywords} \\ \midrule
\multirow{5}{*}{Contradiction} & problem (1.16\%) \\
 & apart (1.15\%) \\
 & also (1.01\%) \\
 & black (0.89\%) \\
 & actually (0.85\%) \\
\midrule
\multirow{5}{*}{Empathy-affiliation} & opinion (1.90\%) \\
 & share (1.64\%) \\
 & understand (1.42\%) \\
 & feel (1.22\%) \\
 & live (1.14\%) \\
\midrule
\multirow{5}{*}{Humor} & hatred (1.40\%) \\
 & solve (1.27\%) \\
 & wow (1.23\%) \\
 & poverty (1.17\%) \\
 & homelessness (1.15\%) \\
\midrule
\multirow{5}{*}{Questions} & comment (1.88\%) \\
 & wonder (1.68\%) \\
 & make (1.60\%) \\
 & facts (1.60\%) \\
 & vague (1.51\%) \\
\midrule
\multirow{5}{*}{Shaming} & others (1.75\%) \\
 & hateful (1.29\%) \\
 & offensive (1.27\%) \\
 & someone (1.17\%) \\
 & without (1.15\%) \\
\midrule
\multirow{5}{*}{Warning of consequences} & online (3.31\%) \\
 & banned (3.07\%) \\
 & permanent (2.22\%) \\
 & choose (1.76\%) \\
 & remember (1.71\%)
   \\ \bottomrule
\end{tabular}
\caption{The table shows the top 5 keywords associated with different types of counterspeech, ranked by their TF-IDF scores. These keywords represent the most distinct and significant terms used within each counterspeech type, reflecting the corresponding discourse's primary themes and focus areas.}
\label{tab:top_keywords}

\end{table}

\section {Definitions}
\label{sec:definitions}

\subsection{Counterspeech type definition}

Here we define the counterspeech types in more details.

\begin{compactitem}
    \item \textbf{Warning of consequences}: Counterspeakers often use this strategy to caution the hate speaker about the potential repercussions of their hate speech. They may remind the speaker of the harm their words can cause to the target group, the lasting impact of online communication, and the possibility of online consequences like reporting and account suspension. This approach highlights the real-world implications of hate speech and can prompt perpetrators to reconsider their words.
    \item \textbf{Shaming/labeling}: Another effective strategy involves labeling hate speech, such as tagging tweets as `hateful', `racist', `bigoted', or `misogynist'. The stigma attached to such labels can prompt individuals to alter their tweets. Counterspeakers also use this strategy to denounce hate speech, helping others identify and respond to it. They may explain to the original speaker why their statement is considered hateful or dangerous, facilitating both condemnation and education.
    \item \textbf{Empathy/affiliation}: This strategy focuses on changing the tone of a hateful conversation. Counterspeakers respond to hostile or hateful messages with a friendly, empathetic, or peaceful tone. They may also establish a connection with the original speaker by affiliating with them or empathising with the group targeted by the hate speech. While the long-term behaviour change is uncertain, this strategy can prevent the escalation of hateful rhetoric and encourage a more constructive exchange.
    \item \textbf{Humor}: Humorous counterspeech is a powerful tool to shift the dynamics of communication, de-escalate conflicts, and draw attention to a message. Counterspeakers may employ humor in various forms, including caricature, sarcasm, and other tones, to neutralize powerful or intimidating hate speech, attract a larger audience, or soften a message that would otherwise be harsh or aggressive.
    \item \textbf{Pointing out hypocrisy}: This strategy involves countering hate speech by pointing out the hypocrisy or contradictions in the user's statements. Counterspeakers may explain and rationalize the hate speaker's previous behaviour or prompt them to resolve to avoid similar behaviour in the future. This approach discredits the accusation and encourages self-reflection.
    \item \textbf{Questions}: Counterspeakers employ this strategy by questioning the sources of information or the rationale behind the hate speaker's claims. By probing and encouraging introspection, this approach can help hate speakers reflect on the content they are promoting, potentially leading to reevaluating their views.
\end{compactitem}

Further, we mention one strategy which should not be used in a typical counterspeech for a given hate speech,  i.e., the annotators should not respond to hateful speech in a hostile, aggressive tone, threat of offline punishment, or insults. This includes but is not limited to the use of profanity, slurs, and name-calling. While annotators should try to counter hate speech, their target should never be to harm the individuals. Finally, we do not use the strategy \textit{facts} as an additional type of counterspeech since factual counterspeech may not be very effective~\cite{benesch2016considerations}. However, we allow the annotators to add any factual information they want to, along with the type mentioned in the task.

\begin{table*}[!htpb]
\centering
\footnotesize
\begin{tabular}{p{1.8cm}|p{1.5cm}|p{11cm}}
\hline
\textbf{Type} & \textbf{Model} & \textbf{Generated sample} \\ 
\hline

\multirow{4}{*}{\shortstack{Warning of\\consequences}} 
& \vspace{1pt} DialoGPT 
& \textbf{Hate speech:} Hey black!  If the system is rigged for whites, then why are the asians doing better than everyone else.  \#MythDestroyed  \#IQMatters
    
  \textbf{Counterspeech:} You shouldn't post things like this, it can come back to you.  Remember, internet is forever, can come back to haunt you. You better choose your words carefully though, the consequences can be quite hurtful. \\
\cline{2-3} 
& \vspace{1pt} Flan-T5 
& \textbf{Hate speech:} "The worst film of 2017" Hopefully worse than that lesbian ghostbusters variant with that fat nigger a few years back

  \textbf{Counterspeech:} This comment is very offensive and can get you banned.\\ 
\cline{2-3} 
&  \vspace{1pt} Llama-2
& \textbf{Hate speech:} In england they call these people asians, even though they are from the sub-continent and pakistan, and east asians are called orientals. It irritates me too, we should classify them by race and perhaps religion. In this case, paki muzzie.
    
  \textbf{Counterspeech:} Using the word "paki" is offensive to the Pakistani people and can lead to a ban on your account. Be careful with your words. \#RespectAllRaces\\ 
\cline{2-3} 
&  \vspace{1pt} Llama-3 
& \textbf{Hate speech:} "Obama didn't do enough for black people!" Not even god can do enough. It's impossible. Maybe if they ship in whites and asians and paint them black?

  \textbf{Counterspeech:} Please do not post such comments. You can be banned for this. The internet is permanent. Don't post something you will regret.\\ 
\hline

\multirow{4}{*}{\shortstack{Shaming}} 
&  \vspace{1pt}DialoGPT 
& \textbf{Hate speech:} I'm not a Racist, I just hate Murderously retarded fascist Ideologies, \& stupid people. But some things are universally recognizable \& some are not, apparently.
    
  \textbf{Counterspeech}: What did the fascist ideologist do to you? Or did you just made a stupid statement on the internet? I doubt it.  Also, you are unlikely to be the victim of any murder or hate speech that is ongoing. \\
\cline{2-3} 
& \vspace{1pt}Flan-T5 
& \textbf{Hate speech}: \#TRUMP 3 yrs ago Obama had 3 million green cards made up.They were 2 go2 the refugees and illegals, DO NOT LET THEM STAY, He also had 5 million SSI cards, which he stole from dead ppl
    
  \textbf{Counterspeech}: Your words hurt people and rapes people. It's bad to think that Obama was such a bad man and he does not have to deal with his immigration problems. \\ 
\cline{2-3} 
&  \vspace{1pt}Llama-2
& \textbf{Hate speech}: Isn't it ironic Freddy, that something as manly as mixed martial arts would be too cuck to allow a 'scary' nationalist on their team. It's run by jews i take it. The ultimate whiny pussies that always get other people to do the fighting for them. heh.
    
  \textbf{Counterspeech}: You are a disgrace. You are a disgrace to the sport and to humanity.\\ 
\cline{2-3} 
&  \vspace{1pt}Llama-3 
& \textbf{Hate speech}: "Sweden decided in 1975 to welcome \#immigration. Since 1975 violent crime in \#Sweden has increased by 300\% and \#RAPE has increased by 1,472\%." Sweden has some of the most beautiful women on the planet and they let them get raped by evil \#muslim \#immigrants. 
    
  \textbf{Counterspeech}: If you don't like it then leave. You're not forced to live there. Besides, the crime rate is a result of the government's inability to handle the problem, not the immigrants.\\ 
\hline

\multirow{4}{*}{\shortstack{Empathy\\Affiliation}} 
&  \vspace{1pt}DialoGPT 
& \textbf{Hate speech}: "Where are all the conservative white men?"  Where are all the WHITE WOMEN defending men from women committing perjury in the divorce courts?  Oh, that's right....they are CHEERING as men commit suicide.  Screw you bitches. Look after yourselves.  \#MGTOW
    
  \textbf{Counterspeech}: We all have our own characteristics, and it's important to work with what we've got, not against. The more you work with it, the more you may become the one with your uniqueness. So it's worth it to explore different perspectives.\\
\cline{2-3} 
& \vspace{1pt}Flan-T5 
& \textbf{Hate speech}: (((They))) are saying there were nigger vikings for fuck sake. Yup jews are rewriting our history to show us that niggers have always been with us. Pisses me off. Fucking jews.
    
  \textbf{Counterspeech}: I think you should learn to appreciate Jewish history better before you start talking about it.\\ 
\cline{2-3} 
&  \vspace{1pt}Llama-2
& \textbf{Hate speech}: "A lawyer representing Jordan Salter, who was accused of bringing a knife to Marjory Stoneman Douglas High School in Parkland, Florida, said he doesn't blame students for wanting to bring in weapons."  Before you see the pic how many would bet it's a niglet caught with a knife?
    
  \textbf{Counterspeech}: I don't think it's right to make generalizations about people based on their race. Everyone has their own story, and we should respect and appreciate diversity. Let's focus on promoting positivity and understanding.\\ 
\cline{2-3} 
&  \vspace{1pt}Llama-3 
& \textbf{Hate speech}: "You (jews) have surpassed all nations in impertinent fables in bad conduct and in barbarism. You deserve to be punished, for this is your destiny." - Voltaire, in a reply to a letter sent to him by a kike complaining about his "anti-semitism."
    
  \textbf{Counterspeech}: I would be more worried about the hate speech than the hate you are accusing others of.  I don't think Voltaire would want you to use his words to spread hate.  Let's all try to be respectful and understanding of each other.\\ 
\hline
\end{tabular}
\caption{Sample generation of counterspeech types (warning-of-consequences, shaming and empathy/affiliation) from each finetuned model.}
\label{tab:gen_counterspeech_types-1-3}
\end{table*}

\begin{table*}[!htpb]
\centering
\footnotesize
\begin{tabular}{p{1.8cm}|p{1.5cm}|p{11cm}}
\hline
\textbf{Type} & \textbf{Model} & \textbf{Generated sample} \\ 
\hline

\multirow{4}{*}{\shortstack{Humor}} 
& \vspace{1pt} DialoGPT 
& \textbf{Hate speech}: \#TRUMP 3 yrs ago Obama had 3 million green cards made up.They were 2 go2 the refugees and illegals,DO NOT LET THEM STAY,He also had 5 million SSI cards,he stole from dead ppl
    
  \textbf{Counterspeech}: I think your lack of knowledge about immigration is showing your ignorance about Trump.   Also you are probably the reason the Trump's popularity is so low.\\
\cline{2-3} 
& \vspace{1pt}Flan-T5 
& \textbf{Hate speech}: 1) strawman  2) strawman  3) blacks voted 88\% democrat in 16, Hispanics and Asians both 65\%, Jewish 71\%, muzzies 62\%.
    
  \textbf{Counterspeech}: It is not that everyone is good at voting, but you might be biased. Not everyone has that opinion, but you are a good person.\\ 
\cline{2-3} 
& \vspace{1pt} Llama-2
& \textbf{Hate speech}: "Obama didn't do enough for black people!" Not even god can do enough. It's impossible. Maybe if they ship in whites and asians and paint them black?
    
  \textbf{Counterspeech}: I'm sure you'd be the first to volunteer to help them paint their faces.  Perhaps you should focus on issues that affect your own community first.  There are plenty of things to fix in your own neighborhood.\\ 
\cline{2-3} 
& \vspace{1pt} Llama-3 
& \textbf{Hate speech}: @user   Cuz it is upsetting all of my muzzie friends fam
    
  \textbf{Counterspeech}: I guess it is good that you have some friends who are Muslims. But I hope you have some non-Muslim friends too. After all, you don't want to be a ghetto with only one race and religion.\\ 
\hline

\multirow{4}{*}{\shortstack{Contradiction}} 
& \vspace{1pt} DialoGPT 
& \textbf{Hate speech}: 2040: Pres. Rchd Spencer \& Vice Pres @user hatch plan 2 get jews \& blacks 2 leave the country. universal basic income creates massive hyperinflation \& the poor \& the jews move! next up 4 our team, getting gays 2 leave (after they've gentrified all the ol' black ghettos:)
    
  \textbf{Counterspeech}: If we are talking about the left, it is worth thinking about the social circle that rejects you outright and you find yourself in.\\
\cline{2-3} 
&\vspace{1pt} Flan-T5 
& \textbf{Hate speech}: 650  uk mps involved in the cover up of 1 million white children raped by paki  muslims. Resign all of you filthy bastards
    
  \textbf{Counterspeech}: And if it's not a crime - it's still a crime to be prosecuted. If it's an infringement - it will be reported.\\ 
\cline{2-3} 
& \vspace{1pt} Llama-2
& \textbf{Hate speech}: Hollywood is out of ideas and is now running out of money so the jews are cannibalizing each other and Weinstein is the first victim. That's my theory anyway.
    
  \textbf{Counterspeech}: There are many successful Hollywood producers and directors who are not Jewish.  And there are many other industries that are not run by Jews.  You are making a generalization based on your own prejudices.\\ 
\cline{2-3} 
& \vspace{1pt} Llama-3 
& \textbf{Hate speech}: It's soooooooo nigger. Future BLM member is what you sound like. Are you from Houston? This sounds like that nigger garbage from that shithole. Damn, break that programming. You are a Jew-tool.
    
  \textbf{Counterspeech}: This is not true. It is not a fact that the black people are from Houston. This is a stereotype. It is not true that the black people are from Houston. It is not true that the black people are garbage.\\ 
\hline

\multirow{4}{*}{\shortstack{Questions}} 
& \vspace{1pt} DialoGPT 
& \textbf{Hate speech}: Stop being obtuse. You know damn well majority of all Mexicans are not white. Hence why you people bitch and moan about white genocide. Which is why u love him.  If it were WHITE EUROPEAN immigrants you wouldn't care neither would trump.   They're not WHITE.
    
  \textbf{Counterspeech}: Why would you use those strong words for? Its not right.  And who would they harm? \\
\cline{2-3} 
& \vspace{1pt} Flan-T5 
& \textbf{Hate speech}: OTOH, you might just be retarded and failing at trolling along with the rest of your life.

  \textbf{Counterspeech}: Are you sure that this is not enough to understand how people treat each other? Do you not want to be perceived as a retarded or not? What would it explain?\\ 
\cline{2-3} 
& \vspace{1pt} Llama-2
& \textbf{Hate speech}: Holy man from Switzerland tells Daily Mail that cancer caused by immigrants.
    
  \textbf{Counterspeech}: How do you know? Do you have any data? What are your sources? Or is this just an opinion?  Do you think immigrants are more likely to get cancer?\\ 
\cline{2-3} 
& \vspace{1pt} Llama-3 
& \textbf{Hate speech}: Judging by your retarded limey accent, I highly doubt your ass has any say regarding American policies. And considering I am an American citizen, you're gona have a hard time making a case against me, old sport.
    
  \textbf{Counterspeech}: What do you have against the British accent? Why do you use such derogatory terms?  Do you have a problem with people who have a different accent?  Do you have a problem with people who are British? \\ 
\hline

\end{tabular}
\caption{Sample generation of counterspeech types (humor, contradiction and questions) from each finetuned model.}
\label{tab:gen_counterspeech_types}
\end{table*}

\section{Top keywords}

The analysis of top keywords for various types of counterspeech reveals distinct themes and focal points within each discourse category. We identify and rank the most significant terms using \texttt{Term Frequency-Inverse Document Frequency (TF-IDF)} scores\footnote{\url{https://scikit-learn.org/stable/modules/generated/sklearn.feature_extraction.text.TfidfVectorizer.html}}. \texttt{TF-IDF}~\cite{Salton1988TermWeightingAI} is a statistical measure that evaluates the importance of a word in a document relative to a collection of documents, where higher scores indicate greater significance within the specific context. We first extract the top keywords for each type and then remove any overlaps to ensure the uniqueness of the terms associated with each category.

Table \ref{tab:top_keywords} showcases the top 5 distinct keywords for different counterspeech types. Understanding these keywords is crucial for identifying the core elements and recurring motifs in counterspeech, which can inform the development of more effective strategies to counteract harmful speech online. For instance, terms like `problem' and  `apart' under the contradiction category indicate a focus on highlighting issues and discrepancies, while keywords such as `opinion. and `share' in empathy-affiliation emphasize the importance of expressing and exchanging personal viewpoints to foster understanding.

\section{Hyperparameters}
\label{sec:hyperparameters}

\subsection{Type classification}

For fine-tuning \texttt{bert-base-cased}, we use a max\_length of 256 and a batch size of 32 with a gradient accumulation steps of 2. We set the learning-rate is 2e-5, number of training epochs of 10 and optimize with \texttt{paged\_adamw\_32bit} having weight decay 0.01. The learning scheduler is set to cosine. We also use an early stopping criteria with a patience of 10 and early stopping threshold of 0.01. For fine-tuning Flan-T5, we use a batch size of 2 with a gradient accumulation steps as 2. We use 10 training epochs along with \texttt{paged\_adamw\_32bit} having weight decay of 0.01. Rest of the things remain same as \texttt{bert-base-uncased}. 

\subsection{Training generation models}
For fine-tuning models, we employ a consistent training configuration across various model types, ensuring both efficiency and performance. The setup includes 5 epochs for Flan-T5 and DialoGPT, and 2 epochs for Llama-2 and Llama-3, with 2 worker processes facilitating efficient data loading in batches of size 2. Gradients are accumulated over 4 steps to manage memory efficiency, and the \texttt{paged\_adamw\_32bit} optimizer is used with a learning rate of 2e-4 and weight decay of 0.001. Mixed precision training with \texttt{fp16} is supported. Logging intervals are set to every 100 steps, with a \texttt{cosine decay} schedule for the learning rate and \texttt{gradient clipping} at a maximum norm of 0.3. The best model is tracked based on 'eval\_loss' and progress is reported to \texttt{Weights \& Biases}\footnote{\url{https://wandb.ai/site}}. Specifically for Llama family models, the \texttt{Low Rank Adaptation (LoRA)}~\cite{hu2021lora} configuration is used along with \texttt{4-bit quantization}~\cite{dettmers2023qlora} which includes hyperparameters such as \texttt{LoRA} $\alpha$ set to 16, dropout rate of 0.1, and a rank of 64, targeting specific model parts. Further \texttt{gradient checkpointing} is used to reduce memory requirements.

\subsection{Generation of responses}
For the Llama family, the generation settings include a batch size of 6, using 4-bit quantization, and a maximum of 50 new tokens. The top-$p$ the sampling parameter is set to 0.9 to control the diversity of the generated output. For Flan-T5 and DialoGPT models, the generation settings differ slightly. The 4-bit quantization parameter is set to false and the batch size is set to 10. The maximum input tokens are fixed based on the particular datasets - Gab (128), Reddit (256) and \sysname{} (128).

\section{System information}

We used the NVIDIA RTX 1080Ti, NVIDIA GTX 2080Ti and NVIDIA Titan Xp having 11-12 GB memory in a Intel(R) Xeon(R) CPU having 32 cores and 250 GB RAM and 128 GB cache. The DialoGPT and FlanT5 models take around 1 hr to train for 5 epochs and Llama family usually takes around 2 hr to train for 2 epochs.

\section{Metrics}

Here, we add some additional details about the metrics that could not be added in the main text. 

\subsection{Evaluation metric considerations}

Here we note some of the choices of metric and their peculiarities. We do not use the BLEU~\cite{papineni2002bleu} score because it has some undesirable properties when used for single sentences, as it is designed to be a corpus-specific measure~\cite{wu2016google}. Further, the reader might notice negative scores in the case of \texttt{bleurt} metric which is not calibrated\footnote{\url{https://github.com/google-research/bleurt/issues/1}}. 

\subsection{Multilabel metrics}
\label{sec:multilabel}
Accuracy is defined as the proportion of predicted \textit{correct} labels to the \textit{total} number of label, averaged over all instances.

\begin{equation}
Accuracy = \frac{1}{\mid D \mid} \displaystyle\sum\limits_{i=1}^{\mid D \mid} \frac{\mid Y_i \cap Z_i \mid }{\mid Y_i \cup Z_i \mid } \label{Accuracy}
\end{equation}

Precision is defined as the proportion of predicted \textit{correct} labels to the total number of \textit{actual} labels, averaged over all instances

\begin{equation}
Precision = \frac{1}{\mid D \mid} \displaystyle\sum\limits_{i=1}^{\mid D \mid} \frac{\mid Y_i \cap Z_i \mid }{\mid Z_i \mid } \label{Precision}
\end{equation}

Recall is defined as the proportion of predicted \textit{correct} labels to the total number of \textit{predicted} labels, averaged over all instances

\begin{equation}
Recall = \frac{1}{\mid D \mid } \displaystyle\sum\limits_{i=1}^{\mid D \mid } \frac{\mid Y_i \cap Z_i \mid}{\mid Y_i \mid} \label{Recall}
\end{equation}

F1-Score is defined simply as the harmonic mean of Precision and Recall.
\begin{equation}
\textit{F1-Score} = 2 * \frac{Precision * Recall}{Precision + Recall} \label{F1}
\end{equation}

Hamming loss  is equal to 1 over
$|D|$ (number  of  multi-label  samples),  multiplied  by  the  sum  of  the symmetric  differences between  the  predictions ($Z_i$)  and  the  true  labels ($Y_i$),  divided  by  the  number  of labels (L),  giving

\begin{equation}
Hamming Loss = \frac{1}{|D|} \displaystyle\sum\limits_{i=1}^{|D|} \frac{|Y_i \Delta Z_i|}{|L|} . \label{HL}
\end{equation}

\section{Prompts}
\label{sec:prompts}
We note the prompts used in this paper which are used for training or zero-shot generation across different models. 

\begin{table*}[!htpb]
    \centering
    \footnotesize
    \begin{tabular}{p{3cm}p{3cm}p{8cm}}
    \toprule
    \textbf{Task} & \textbf{Model(s)} & \textbf{Prompt}\\\midrule
    \multirow{3}{*}{Vanilla CS gen} & Flan-T5 and DialoGPT & Counterspeech is a strategic response to hate speech, aiming to foster understanding or discourage harmful behavior. A good counterspeech to this hate speech - \textbf{"\{hate\_speech\}"} is:\\\cline{2-3}
    & Llama-2 &  [INST] <<SYS>>  You are an helpful agent who generates a specific type of counterspeech to the hate speech provided by the user. Definition: Counterspeech is a strategic response to hate speech, aiming to foster understanding or discourage harmful behavior. <</SYS>> \textbf{\{hate\_speech\}} [/INST]\\\cline{2-3}
    & Llama-3 & <|begin\_of\_text|><|start\_header\_id|>system <|end\_header\_id|> You are an helpful agent who generates a specific type of counterspeech to the hate speech provided by the user. Definition: Counterspeech is a strategic response to hate speech, aiming to foster understanding or discourage harmful behavior.<|eot\_id|> <|start\_header\_id|> user <|end\_header\_id|> \textbf{\{hate\_speech\}} <|eot\_id|><|start\_header\_id|> assistant <|end\_header\_id|> \\
    \midrule
    \multirow{3}{*}{Type-spec CS gen} & Flan-T5 and DialoGPT & Counterspeech is a strategic response to hate speech, aiming to foster understanding or discourage harmful behavior. Different types of counterspeech include: \{Definitions of different counterspeech\}. A \textbf{"\{type\}"} type good counterspeech to this hate speech -\textbf{\{hate\_speech\}} is:\\\cline{2-3}
    & Llama-2 & <|begin\_of\_text|><|start\_header\_id|>system<|end\_header\_id|>You are an helpful agent who generates a counterspeech of type - \textbf{\{type\}} to the hate speech provided by the user. Definition: Counterspeech is a strategic response to hate speech, aiming to foster understanding or discourage harmful behavior. Different types of counterspeech include: \textbf{\{Definitions of different counterspeech\}} <|eot\_id|><|start\_header\_id|>user<|end\_header\_id|> \textbf{{hate\_speech}} <|eot\_id|><|start\_header\_id|>assistant <|end\_header\_id|>\\\cline{2-3}
    & Llama-3 & [INST] <<<SYS>>  You are an helpful agent who generates a counterspeech of type - \textbf{\{type\}} to the hate speech provided by the user. Definition: Counterspeech is a strategic response to hate speech, aiming to foster understanding or discourage harmful behavior. Different types of counterspeech include: \textbf{\{Definitions of different counterspeech\}} <</SYS>>\textbf{\{hate\_speech\}} [/INST]\\
    \midrule
    \multirow{3}{*}{CS-Type} & Flan-T5 and GPT-4 & Counterspeech is a strategic response to hate speech, aiming to foster understanding or discourage harmful behavior. Different types of counterspeech include: \textbf{\{Definitions of different counterspeech\}}. Given this counterspeech - \textbf{\{counterspeech\}} what are the types present in the counterspeech out of the ones listed ? Give in the format of a list\\
   \bottomrule
    \end{tabular}
    \caption{This table notes down the prompts used for different models in zero-shot/ training pipelines. We show prompts for Vanilla Counterspeech Generation (Vanilla CS Gen), Type specific Counterspeech Generation (Type-spec CS Gen) and Counter speech type classification (CS-Type).}
    \label{tab:prompts}
\end{table*}

\end{document}